\documentclass[10pt,twocolumn,letterpaper]{article}

\usepackage[pagenumbers]{cvpr} 

\usepackage{graphicx}
\usepackage{amsmath}
\usepackage{amssymb}
\usepackage{booktabs}
\usepackage{multirow}
\usepackage{comment}
\usepackage{caption}
\usepackage{wrapfig}
\usepackage{enumitem}
\usepackage{subcaption}

\usepackage{array}
\usepackage{color}
\usepackage{algorithm}
\usepackage{colortbl}
\definecolor{greyline}{rgb}{0.105,0.410,0.113}

\newcommand{\squishlist}{
	\begin{list}{$\bullet$}
		{ \setlength{\itemsep}{0pt}
			\setlength{\parsep}{1pt}
			\setlength{\topsep}{1pt}
			\setlength{\partopsep}{0pt}
			\setlength{\leftmargin}{1.5em}
			\setlength{\labelwidth}{1em}
			\setlength{\labelsep}{0.5em} } }
\newcommand{\squishend}{\end{list} 
}

\usepackage[pagebackref,breaklinks,colorlinks]{hyperref}

\usepackage[capitalize]{cleveref}
\crefname{section}{Sec.}{Secs.}
\Crefname{section}{Section}{Sections}
\Crefname{table}{Table}{Tables}
\crefname{table}{Tab.}{Tabs.}

\begin{document}

\title{RegionCLIP: Region-based Language-Image Pretraining}

\author{
Yiwu Zhong$^{1}$\thanks{Work done as an intern at Microsoft Research.}, Jianwei Yang$^{2}$, Pengchuan Zhang$^{2}$, Chunyuan Li$^{2}$, Noel Codella$^{3}$,\\ Liunian Harold Li$^{4}$, 
Luowei Zhou$^{3}$, Xiyang Dai$^{3}$, Lu Yuan$^{3}$, Yin Li$^{1}$, Jianfeng Gao$^{2}$ \\
$^1$University of Wisconsin-Madison, $^2$Microsoft Research, $^3$Microsoft Cloud + AI, $^4$UCLA\\
\texttt{\scriptsize{\{yzhong52, yin.li\}@wisc.edu,\{jianwei.yang, penzhan, chunyl, ncodella,}} \\ 
\texttt{\scriptsize{luozhou, xidai, luyuan, jfgao\}@microsoft.com, \{liunian.harold.li\}@cs.ucla.edu}}\\

}

\maketitle

\begin{abstract}
Contrastive language-image pretraining (CLIP) using image-text pairs has achieved impressive results on image classification in both zero-shot and transfer learning settings. However, we show that directly applying such models to recognize image regions for object detection leads to poor performance due to a domain shift: CLIP was trained to match an image as a whole to a text description, without capturing the fine-grained alignment between image regions and text spans. To mitigate this issue, we propose a new method called RegionCLIP that significantly extends CLIP to learn region-level visual representations, thus enabling fine-grained alignment between image regions and textual concepts. Our method leverages a CLIP model to match image regions with template captions, and then pretrains our model to align these region-text pairs in the feature space. When transferring our pretrained model to the open-vocabulary object detection task, our method outperforms the state of the art by \textbf{3.8 AP50} and \textbf{2.2 AP} for novel categories on COCO and LVIS datasets, respectively. Further, the learned region representations support zero-shot inference for object detection, showing promising results on both COCO and LVIS datasets. Our code is available at \url{https://github.com/microsoft/RegionCLIP}.

\end{abstract}

\section{Introduction}
\label{sec:intro}

\begin{figure}[t]
	\centering
	\includegraphics[width=0.99\linewidth]{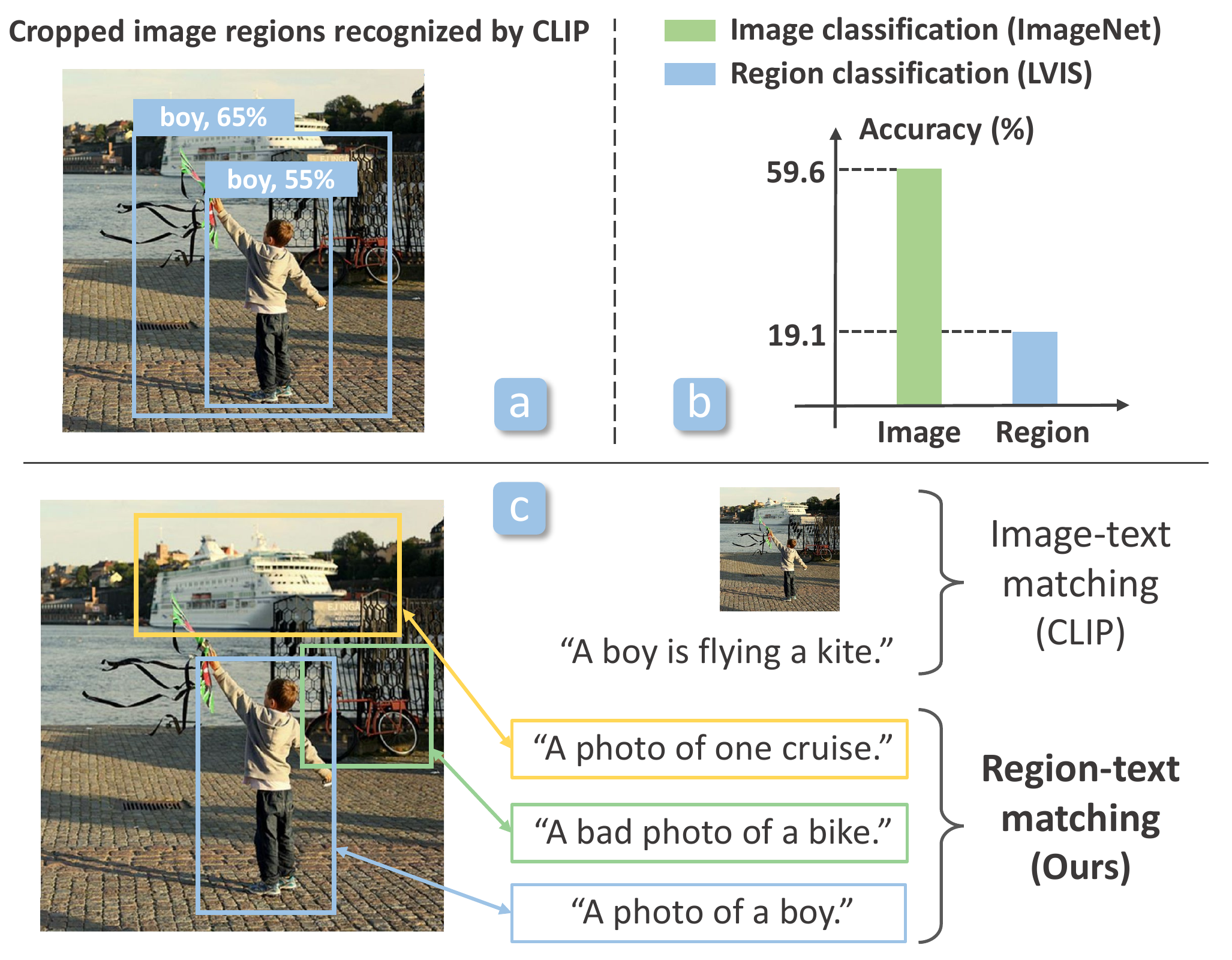} \vspace{-0.5em}
    \caption{{\bf (a)}.\ A pretrained CLIP model~\cite{radford2021learning} failed to capture localization quality. {\bf (b)}.\ A major drop on accuracy when using the same pretrained CLIP to classify image regions. {\bf (c)}.\ Our key idea is learning to match {\it image regions} and their text descriptions.}
    \label{fig:teaser} 
\end{figure}

Recent advances in vision-language representation learning has created remarkable models like CLIP~\cite{radford2021learning} and ALIGN~\cite{jia2021scaling}. Such models are trained using hundreds of millions of image-text pairs by matching images to their captions, achieving impressive results of recognizing a large set of concepts without manual labels, and capable of transferring to many visual recognition tasks. Following their success on image classification, a natural question is that whether these models can be used to reason about image regions, \eg, for tasks like object detection. 

To answer this question, we construct a simple R-CNN style~\cite{girshick2014rich} object detector using a pretrained CLIP model, similar to adapting a pretrained convolutional network. This detector crops candidate object regions from an input image, and applies the CLIP model for detection by matching visual features of cropped regions to text embeddings of object categories. Fig.\ \ref{fig:teaser}(a-b) shows the results on LVIS dataset~\cite{gupta2019lvis}. When using object proposals~\cite{faster-rcnn} as the input regions, scores from CLIP often fail to capture the localization quality (Fig.\ \ref{fig:teaser}a). Even with ground-truth object boxes, classification accuracy using CLIP drops significantly from 60\% on ImageNet to 19\% on LVIS, with a similar number of classes (Fig.\ \ref{fig:teaser}b). There is thus a major performance degradation when applying a pretrained CLIP model for object detection. {\it How can we empower a vision-language pretrained model to reason about image regions?}

We believe the main gap lies in the training of these vision-language models. Many existing vision-language models, including CLIP, are trained to match an image with its image-level text description. The training is unaware of the alignment between local image regions and text tokens. Thus, the models are unable to precisely ground a textual concept to an image region. Further, cropping image regions and matching them to text tokens largely ignore the surrounding visual context that is critical for object recognition, not to mention the high computational cost, \eg a few seconds per image on a modern GPU. 

In this paper, we explore learning {\it region representations} for object detection via vision-language pretraining. Our key idea
is to explicitly align image regions and text tokens during pretraining. 
However, two key challenges arise. 
First, the fine-grained alignment between image regions and text tokens is not available in image-text pairs.
Second, the text description of its paired image is often incomplete, \ie many image regions are not described by the text. 
To address these challenges, we propose to bootstrap from a pretrained vision-language model to align image regions and text tokens, and to fill in the missing region descriptions, as illustrated in Fig.\ \ref{fig:teaser}c. 

Specifically, our method starts with a pool of object concepts parsed from text corpus, and synthesizes region descriptions by filling these concepts into pre-defined templates. Given an input image and its candidate regions from either object proposals or dense sliding windows, a pretrained CLIP model is used to align the region descriptions and the image regions, creating ``pseudo'' labels for region-text alignment. Further, we use both ``pseudo'' region-text pairs and ground-truth image-text pairs to pretrain our vision-language model via contrastive learning and knowledge distillation. Although the ``pseudo'' region-text pairs are noisy, they still provide useful information for learning region representations and thus bridge the gap to object detection, as validated by our experiments.

We pretrain our models on captioning datasets (\eg, Conceptual Caption) and mainly evaluate models on the benchmarks of open-vocabulary object detection (COCO and LVIS datasets). 
When transferred to open-vocabulary object detection, our pretrained model establishes new state-of-the-art (SoTA) results on COCO and LVIS. For instance, our model achieves a relative gain of \textbf{37.7}\% over published SoTA in AP50 for novel categories on COCO. Moreover, our model supports zero-shot inference and outperforms baselines by a clear margin. 

Our contributions are summarized as follows:
(1) We propose a novel method that aligns image regions and their descriptions without manual annotation, thereby enabling vision-language pretraining for learning visual region representations.
(2) A key technical innovation that facilitates our pretraining is a scalable approach for generating region descriptions, neither relying on human annotations nor limited to the text paired with an image.
(3) Our pretrained model presents strong results when transferred to open-vocabulary object detection, and demonstrates promising capability on zero-shot inference for object detection.

\section{Related Work}
\label{sec:related_work}

\noindent \textbf{Visual representation learning for images}. 
Early works on visual representation learning focused on learning from intensive human labels by training image classifiers~\cite{NIPS2012_c399862d,simonyan2014very,szegedy2015going,he2016deep,dosovitskiy2020image}. These classifiers can be further used to label un-annotated images for training student models in semi-supervised learning~\cite{pham2021meta,xie2020self,yalniz2019billion}.
To reduce the annotation burden, self-supervised learning~\cite{he2020momentum,chen2020simple,grill2020bootstrap,caron2020unsupervised} was proposed to match the visual representation of different views from the same image. 
The most relevant work is learning from natural language, such as image tags~\cite{Hironobu99imagetoword,matchwordpicture,divvala2014learning,chen2015webly,joulin2016learning} and text descriptions~\cite{wang2009learning,he2017fine,sariyildiz2020learning,desai2020virtex,zhong2021learning}. Coupled with millions of image-text pairs collected from the Internet, recent vision-language pretraining~\cite{radford2021learning,jia2021scaling} learned to match images with image descriptions and demonstrated impressive performance on zero-shot inference and transfer learning for image classification. 
However, these works focus on image representation and target at image classification. In this paper, we propose to learn visual representation for image regions which supports zero-shot inference and transfer learning for region reasoning tasks (\eg, object detection).

\noindent \textbf{Visual representation learning for image regions}. By leveraging human annotations contributed by~\cite{pascal_voc,lin2014microsoft,krishna2017visual,gupta2019lvis}, major progress has been made to reason about image regions, such as object detection~\cite{faster-rcnn,redmon2016you,tian2019fcos,carion2020end}. With the object detectors trained on these human annotations as teacher models, semi-supervised learning~\cite{Xu_2021_ICCV,zoph2020rethinking,sohn2020simple} creates pseudo labels for image regions in return for training student detectors. Beyond object labels, the region representation learned from additional labels of object attributes~\cite{krishna2017visual,Anderson2017up-down,Zhang_2021_CVPR} demonstrated noticeable improvement on vision-language tasks~\cite{yu2021ernie,chen2020uniter,li2020oscar,tan2019lxmert,zhou2020unified,lu2019vilbert}. However, these works heavily rely on expensive human annotation and are limited to predefined categories. To reduce annotation cost, the idea of self-supervised learning is extended to region representation learning~\cite{Ramanathan_2021_ICCV, Henaff2021ICCV} by maximizing the representation similarity among augmented views of image regions.
Different from these works, we propose to learn region representation via vision-language pretraining, inspired by CLIP~\cite{radford2021learning}. The learned region representation supports recognizing image regions with a large vocabulary. 

\begin{figure*}
	\centering
	\includegraphics[width=0.95\linewidth]{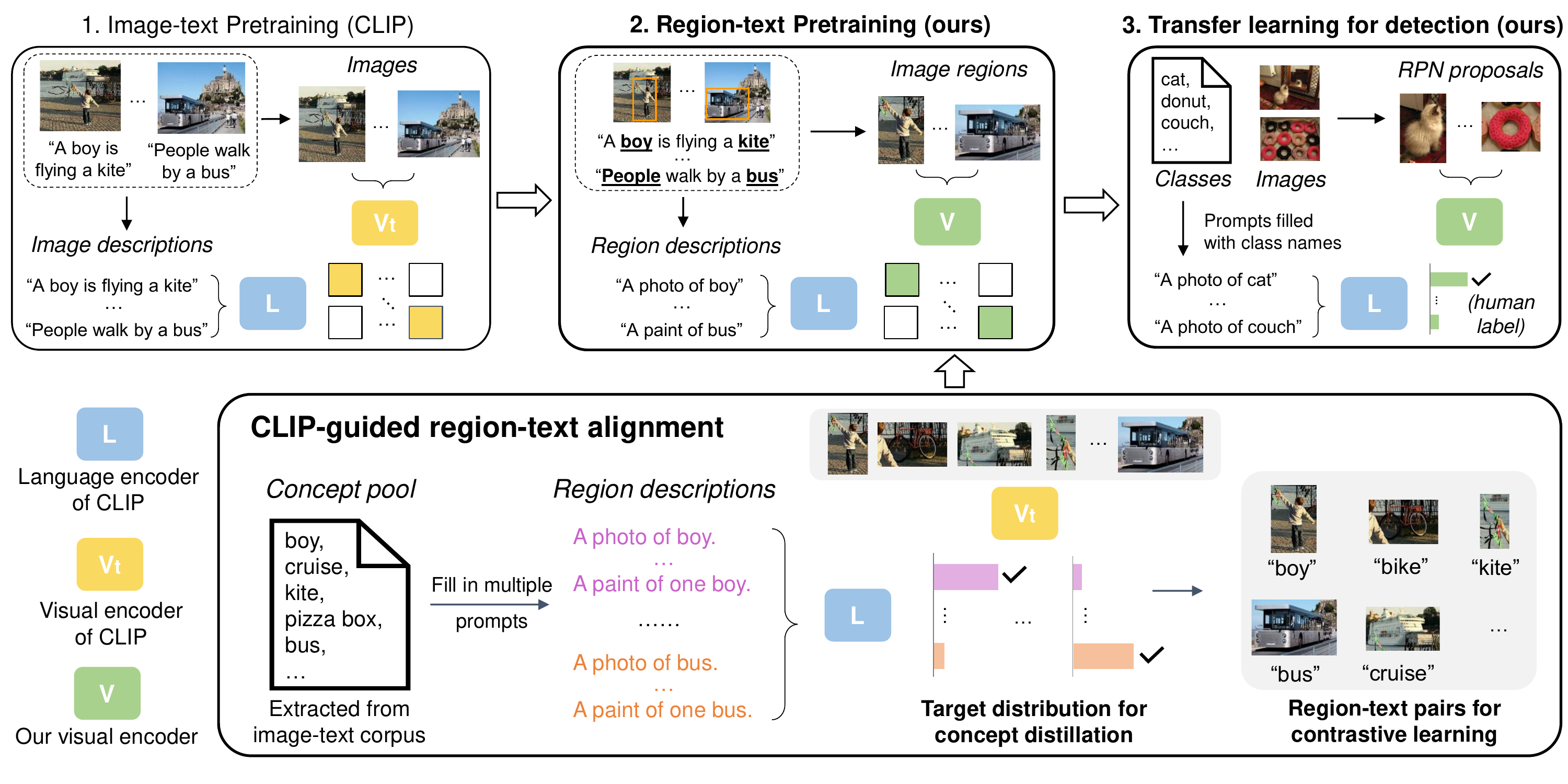} \vspace{-1.0em}
    \caption{Method overview. We propose to learn visual representation for image regions via vision-language pretraining. Panel 1: With contrastive learning, CLIP is able to match images and their descriptions. Panel 2: Initialized by pretrained CLIP, our visual encoder learns visual region representation from the created region-text pairs. Specifically, as shown in the bottom row, we first create texts by filling the prompts with object concepts which are parsed from image descriptions, then use pretrained CLIP to align these texts and image regions proposed by RPN. Panel 3: When human annotation for image regions is available, we transfer our visual encoder for object detection.}
    \label{fig:model_overview} 
\end{figure*}

\noindent \textbf{Zero-shot and open-vocabulary object detection}. Zero-shot object detection aims at detecting novel object classes which are not seen during detector training~\cite{bansal2018zero,rahman2020zero,zareian2021open,gu2021zero,Zhu_2020_CVPR,Rahman_aaai2020}. Bansal \etal~\cite{bansal2018zero} learned to match the visual features of cropped image regions to word embeddings using max-margin loss. 
Rahman \etal~\cite{Rahman_aaai2020} proposed polarity loss to model background category and to cluster categories with similar semantics. Zhu \etal~\cite{Zhu_2020_CVPR} explored improving localization performance for novel categories by synthesizing visual features with a generative model. These zero-shot object detectors usually rely on the semantic space of pretrained word embeddings~\cite{pennington2014glove}. 
Recently, Zareian \etal~\cite{zareian2021open} proposed OVR for open-vocabulary object detection, where a visual encoder was first pretrained on image-text pairs to learn broad object concepts and then transferred to zero-shot object detection setting. Another close work is ViLD~\cite{gu2021zero} that focused on the training of zero-shot object detectors by distilling visual features from a pretrained CLIP model~\cite{radford2021learning}. Similar to OVR and ViLD, our detector also leverages the visual-semantic space learned from vision-language pretraining. Different from OVR, we propose to learn visual region representation from our ``pseudo'' region-text pairs given by another pretrained CLIP model. Our method is thus not restricted to particular text that pairs with an image. Unlike ViLD, our method focuses on pretraining and the resulting regional representations support both zero-shot inference and transfer learning.

\section{Method}
\label{sec:method}

\subsection{Problem Definition}

Our goal is to learn a regional visual-semantic space which covers rich object concepts so that it can be used for open-vocabulary object detection. Consider a text description $t$ that describes the content of region $r$ in an image $I$. In the visual-semantic space, the visual region representation $\mathcal{V}(I,r)$ extracted from $r$ should be matched to text representation $\mathcal{L}(t)$. $\mathcal{V}$ is a visual encoder that takes image $I$ and a region location $r$, and outputs a visual representation for this region. $\mathcal{L}$ is a language encoder that converts a text in natural language to a semantic representation. 

\smallskip
\noindent \textbf{Disentanglement of recognition and localization}. There are two key components for image region understanding: localization and recognition. Inspired by~\cite{singh2018r}, we disentangle these two components, use existing region localizers, and focus on region recognition by learning regional visual-semantic space without heavy human annotation.

\smallskip
\noindent \textbf{Method overview.}  
As shown in Fig.\ \ref{fig:model_overview}, we denote $\mathcal{V}_t$ and $\mathcal{L}$ as visual and language encoders pretrained to match images to their descriptions, such as CLIP. Our goal is to train a visual encoder $\mathcal{V}$ so that it can encode image regions and match them to region descriptions encoded by language encoder $\mathcal{L}$. To address the challenge of lacking large-scale region descriptions, as shown at the bottom of Fig.\ \ref{fig:model_overview}, we construct a pool of object concepts, create the region descriptions by filling concepts into prompts, and leverage teacher encoder $\mathcal{V}_t$ to align these text descriptions with the image regions proposed by an image region localizer. Given the created region-text pairs, our visual encoder $\mathcal{V}$ learns to match these pairs via contrastive learning and concept distillation. Once pretrained, our model supports zero-shot inference for region recognition and can be transferred to object detector when the human annotation is available.

\subsection{Region-based Language-Image Pretraining}
\label{sec:pretraining}
We introduce how we obtain region-level visual and semantic representation, and then describe how we build the alignment between image regions and region descriptions. 

\subsubsection{Visual and Semantic Region Representation}
\noindent \textbf{Visual region representation.}
Image regions can be proposed by either off-the-shelf object localizers (\eg, RPN~\cite{faster-rcnn}) or dense sliding windows (\eg, random regions). By default, we use an RPN which is pretrained on human-annotated object bounding boxes \emph{without} object labels. We use RPN to propose image regions for all images in a batch and finally obtain $N$ image regions in total. 
The set of image regions denotes as $\{r_i\}_{i=1,...,N}$.
Given the proposed regions, the visual representation $v_i$ of region $r_i$ is extracted from our visual encoder $\mathcal{V}$ with a feature pooling method, such as RoIAlign~\cite{he2017mask}. RoIAlign pools regional visual features from the feature map of full image by using interpolation. 
Specially, we note that our visual encoder $\mathcal{V}$ is initialized by the teacher $\mathcal{V}_t$ so that it can have a good starting point in visual-semantic space. 

\smallskip
\noindent \textbf{Semantic region representation.}
A single image usually contains rich semantics, covering one or more objects out of thousands of categories. It is costly to annotate all these categories in the large-scale image-text datasets. To this end, we first build a large pool of concepts to exhaustively cover regional concepts, regardless of individual full images. 
As shown at the bottom of Fig.\ \ref{fig:model_overview}, 
we create a pool of object concepts which are parsed from text corpus (\eg, the image descriptions collected from the Internet), by using off-the-shelf language parsers~\cite{sgparser,schuster-etal-2015-generating}. 
Given the concept pool, the semantic representations for regions are created by two steps: (1) We create a short sentence for each concept by filling it to prompt templates (\eg, prompts of CLIP~\cite{radford2021learning}). For example, the ``kite'' concept will be converted to ``A photo of a {\it kite}''. (2) We encode the created text descriptions into semantic representations by using the pretrained language encoder $\mathcal{L}$. 
Finally, all regional concepts are represented by their semantic embeddings $\{l_j\}_{j=1,...,C}$ and $C$ denotes the size of concept pool. 

While our region descriptions are built based on the image descriptions, our method is not constrained by the particular text description that pairs with an image. More importantly, in light of the powerful language encoder $\mathcal{L}$ which has seen many words in natural language, we can easily customize our concept pool and scale it up, which is difficult to achieve from human annotations. Similarly, in vision modality, the disentanglement of visual recognition and localization makes our method flexible to adopt different ways of extracting candidate regions.

\subsubsection{Visual-Semantic Alignment for Regions}

\noindent\textbf{Alignment of region-text pairs.} We leverage a teacher visual encoder $\mathcal{V}_t$ to create the correspondence between image regions and our created texts (represented as semantic embeddings). Again, visual representation $v^{t}_{i}$ of region $r_i$ is extracted from teacher encoder $\mathcal{V}_t$ by pooling features from the loca image region with RoIAlign. Then we compute the matching score between $v^{t}_{i}$ and each concept embedding $l_j$. The matching score $S(v, l)$ is given by
\begin{equation}
\small
S(v, l) = \frac {v^{T} \cdot l } {||v||\cdot||l||}.
\end{equation}

The object concept $l_m$ that has highest matching score is selected and linked to region $r_i$. Finally, we obtain the pseudo labels for each region, namely the pairs of $\{v_i,l_m\}$.

\smallskip
\noindent\textbf{Our pretraining scheme.} Our pretraining leverages both created region-text pairs and the image-text pairs from the Internet. Given the aligned region-text pairs (represented by $\{v_i,l_m\}$), we pretrain our visual encoder with contrastive loss and distillation loss based on the image regions across different images. The contrastive loss is computed as
\begin{equation}
\small
L_{cntrst} = \frac{1}{N} \sum_{i} -\log(p(v_i,l_m)),
\end{equation}
where $p(v_i,l_m)$ is given by
\begin{equation}
\resizebox{0.85\hsize}{!}{%
$p(v_i,l_m) = \frac {\exp(S(v_i, l_m)/\tau)} {\exp(S(v_i, l_m)/\tau) + \sum_{k\in \mathcal{N}_{r_i}} \exp(S(v_i, l_k)/\tau)}
$.
}
\end{equation}
$\tau$ is a predefined temperature. $\mathcal{N}_{r_i}$ represents a set of negative textual samples for region $r_i$, \ie, the object concepts that are not matched to region $r_i$ but matched to other regions in the batch. Beyond contrastive learning over positive and negative region-text pairs, we also consider knowledge distillation for each image region over all object concepts. 
The distillation loss is defined as
\begin{equation}
\small
L_{dist} = \frac{1}{N} \sum_{i} L_{KL}(q^t_{i}, q_i),
\end{equation}
where $L_{KL}$ is KL divergence loss; both $q^t_{i}$ and $q_i$ are probabilities over all object concepts. $q^t_{i}$ is a soft target from teacher model computed as $softmax(S(v^{t}_{i},l_1)/\tau,...,S(v^{t}_{i},l_C)/\tau)$. $q_i$ is computed as $softmax(S(v_i,l_1)/\tau,...,S(v_i,l_C)/\tau)$ coming from our student model. 

Given image-text pairs collected from the Internet, our region-level contrastive loss $L_{cntrst}$ can naturally extend to image-level contrastive loss $L_{cntrst-img}$. It can be considered as a special case where (1) the visual representation is extracted for single global box that covers the whole image, (2) text descriptions are collected from the Internet, and (3) negative samples are the text descriptions that come with other images. Finally, our overall loss function is given by
\begin{equation}
\small
L = L_{cntrst} + L_{dist} + L_{cntrst-img}.
\end{equation}

\noindent\textbf{Zero-shot inference}. Once pretrained, our visual encoder can be directly applied to region reasoning tasks. For example, given image region proposals from RPN, region representation extracted from our visual encoder are matched to the embeddings of target object concepts, thereby predicting the most likely category. 
Inspired by~\cite{singh2018r,zhou2021probabilistic}, we fuse RPN objectness scores and category confidence scores by geometry mean. Empirically, we observe that RPN scores significant improve zero-shot inference.

\subsection{Transfer Learning for Object Detection}
\label{sec:transfer}

In pretraining, our visual encoder learns from region-text alignment which is created by teacher model. Such alignment does not require human efforts but it is inevitably noisy and weak. When strong supervision for image regions is available (\eg, the human-annotated detection labels), our visual encoder can be further fine-tuned by simply replacing the region descriptions, as shown in Panel 3 of Fig.\ \ref{fig:model_overview}.

Specifically, we transfer our pretrained visual encoder to object detectors by initializing their visual backbones. To detect image objects, same as our pretraining, we use an off-the-shelf RPN to localize object regions and recognize these regions by matching their visual region representation and the semantic embeddings of target object classes (\eg, the object classes in detection dataset).

\smallskip
\noindent\textbf{Training for open-vocabulary object detection}~\cite{zareian2021open}. In this setting, the detectors are trained by the annotation of base categories while expected to detect novel categories never seen in detector training. Specially, we apply class-wise weighted cross-entropy loss to train our detectors.
(1) For base categories, inspired by focal loss~\cite{lin2017focal}, we apply focal scaling and calculate the weight for a base category as $(1 - p^b)^\gamma$, where $p^b$ is probability after softmax for this base category and $\gamma$ is a hyperparameter. Empirically, focal scaling is effective to alleviate the forgetting of previously learned object concepts in pretraining, especially when there are very few base categories in dataset (\eg, COCO). We conjecture that the detector might overfit to the small set of base categories, thereby hurting the generalization on novel categories.
(2) For background category, we use a fixed all-zero embedding and apply a predefined weight to background regions following~\cite{zareian2021open}.

\section{Experiments}
\label{sec:experiments}

Our models are primarily evaluated on transfer learning for open-vocabulary object detection. We also present results of zero-shot inference for object detection. Finally, we present ablation study on different model components.

\smallskip
\noindent \textbf{Datasets}. For pretraining, we use the image-text pairs from Conceptual Caption dataset (CC3M)~\cite{sharma2018conceptual} which collects 3 millions of image-text pairs from the web. We also consider a smaller dataset COCO Caption (COCO Cap)~\cite{chen2015microsoft} to pretrain our model when conducting ablation study. COCO Cap contains 118k images with each image annotated by human for 5 captions. We parsed object concepts from COCO/CC3M dataset and filtered the concepts whose frequency is lower than 100, resulting in 4764/6790 concepts. 

For transfer learning of open-vocabulary object detection, we train detectors with base categories of COCO detection dataset~\cite{lin2014microsoft} and LVIS dataset (v1)~\cite{gupta2019lvis}, respectively. On COCO, We follow the data split of \cite{bansal2018zero} with 48 base categories and 17 novel categories which are subsets of COCO object classes. We use the processed data from \cite{zareian2021open} with 107,761 training images and 4,836 test images. On LVIS, following \cite{gu2021zero}, we use the training/validation images for training/evaluation and adopt the category split with 866 base categories (common and frequent objects) and 337 novel categories (rare objects). 

We evaluate object detection performance on COCO and LVIS for both transfer learning and zero-shot inference.

\begin{table*}[]
\centering
\resizebox{0.73\textwidth}{!}{%
\begin{tabular}{lll|ll|ccccc}
\toprule
\multicolumn{3}{c|}{\multirow{2}{*}{\begin{tabular}[c]{@{}c@{}}Visual Encoder Pretraining\end{tabular}}} & \multicolumn{2}{c|}{\multirow{2}{*}{\begin{tabular}[c]{@{}c@{}}Detector Training \end{tabular}}} & \multicolumn{5}{c}{COCO} \\ 
\multicolumn{3}{c|}{} & \multicolumn{2}{c|}{} & \multicolumn{1}{c}{\multirow{2}{*}{\begin{tabular}[c]{@{}c@{}}Novel \\ (17)\end{tabular}}} & \multicolumn{1}{c}{\multirow{2}{*}{\begin{tabular}[c]{@{}c@{}}Base \\ (48)\end{tabular}}} & \multicolumn{3}{c}{Generalized (17+48)} \\ 
Method & Dataset & Backbone & Method & Backbone & \multicolumn{1}{c}{} & \multicolumn{1}{c}{} & Novel & Base & All \\ \midrule
Cls-ResNet~\cite{he2016deep} & ImageNet & RN50 & FR-CNN~\cite{faster-rcnn} & RN50-C4 & \multicolumn{1}{c}{-} & \multicolumn{1}{c}{54.5} & - & - & - \\
Cls-IncRN~\cite{szegedy2017inception} & ImageNet & IncRNv2 & SB~\cite{bansal2018zero} & IncRNv2 & \multicolumn{1}{c}{0.70} & \multicolumn{1}{c}{29.7} & 0.31 & 29.2 & 24.9 \\
Cls-DarkNet~\cite{redmon2016you} & ImageNet & DarkNet19 & DELO~\cite{Zhu_2020_CVPR} & DarkNet19 & \multicolumn{1}{c}{7.60} & \multicolumn{1}{c}{14.0} & 3.41 & 13.8 & 13.0 \\
Cls-ResNet~\cite{he2016deep} & ImageNet & RN50 & PL~\cite{Rahman_aaai2020} & RN50-FPN & \multicolumn{1}{c}{10.0} & \multicolumn{1}{c}{36.8} & 4.12 & 35.9 & 27.9 \\
OVR~\cite{zareian2021open} & COCO Cap & RN50 & OVR~\cite{zareian2021open} & RN50-C4 & \multicolumn{1}{c}{27.5} & \multicolumn{1}{c}{46.8} & 22.8 & 46.0 & 39.9 \\
OVR~\cite{zareian2021open} & CC3M & RN50 & OVR~\cite{zareian2021open} & RN50-C4 & \multicolumn{1}{c}{16.7} & \multicolumn{1}{c}{43.0} & - & - & 34.3 \\
CLIP~\cite{radford2021learning} & CLIP400M & ViT-B/32 & ViLD*~\cite{gu2021zero} & RN50-FPN & \multicolumn{1}{c}{-} & \multicolumn{1}{c}{-} & 27.6 & \textbf{59.5} & \textbf{51.3} \\ \midrule
CLIP~\cite{radford2021learning} & CLIP400M & RN50 & Ours & RN50-C4 & \multicolumn{1}{c}{22.5} & \multicolumn{1}{c}{53.1} & 14.2 & 52.8 & 42.7 \\
Ours & COCO Cap & RN50 & Ours & RN50-C4 & \multicolumn{1}{c}{30.8} & \multicolumn{1}{c}{55.2} & 26.8 & 54.8 & 47.5 \\
Ours & CC3M & RN50 & Ours & RN50-C4 & \multicolumn{1}{c}{\textbf{35.2}} & \multicolumn{1}{c}{\textbf{57.6}} & \textbf{31.4} & 57.1 & 50.4 \\ \midrule
Ours & CC3M & RN50x4 & Ours & RN50x4-C4 & \multicolumn{1}{c}{\textbf{43.3}} & \multicolumn{1}{c}{\textbf{61.9}} & \textbf{39.3} & \textbf{61.6} & \textbf{55.7} \\ \bottomrule
\end{tabular}%
} 
\vspace{-2mm}
\caption{Open-vocabulary object detection results on COCO dataset. Initialized by our pretrained visual encoder, our detector outperforms published works on all metrics by a remarkable margin, and outperforms the unpublished work ViLD* on novel categories. ViLD* trains the detector with data augmentation of copy-paste~\cite{ghiasi2021simple} and a much longer training schedule (16x). Notations: Cls denotes the image classification pretraining on ImageNet~\cite{imagenet}, RN50 means ResNet50, IncRNv2 is Inception-ResNet-V2.}
\label{tab:zeroshot-main}
\end{table*}

\begin{table*}[]
\centering
\resizebox{0.92\textwidth}{!}{%
\begin{tabular}{lll|llll|llll}
\toprule
\multicolumn{3}{c|}{Visual Encoder Pretraining} & \multicolumn{4}{c|}{Detector Training} & \multicolumn{4}{c}{LVIS} \\ 
Method & Dataset & Backbone & Method & Backbone & Training Strategy & Supervision & APr & APc & APf & mAP \\ \midrule
- & - & - & Mask RCNN~\cite{he2017mask} & RN50-FPN & 16x+Copy-paste~\cite{ghiasi2021simple} & Base+Novel & 13.0 & 26.7 & \textbf{37.4} & 28.5 \\
Cls-ResNet~\cite{he2016deep} & ImageNet & RN50 & Mask RCNN~\cite{he2017mask} & RN50-C4 & 1x+Standard & Base+Novel & 11.9 & 22.0 & 29.7 & 23.3 \\ \midrule
CLIP~\cite{radford2021learning} & CLIP400M & ViT-B/32 & ViLD*~\cite{gu2021zero} & RN50-FPN & 16x+Copy-paste~\cite{ghiasi2021simple} & Base & 16.7
& 26.5
& 34.2
& 27.8
\\
Ours & CC3M & RN50 & Ours & RN50-C4 & 1x+Standard & Base & 17.1
& 27.4
& 34.0
& 28.2
\\ \midrule
CLIP~\cite{radford2021learning} & CLIP400M & ViT-B/32 & ViLD*~\cite{gu2021zero} & RN152-FPN & 16x+Copy-paste~\cite{ghiasi2021simple} & Base & 19.8 & 27.1 & 34.5  & 28.7 \\ 
Ours & CC3M & RN50x4 & Ours & RN50x4-C4 & 1x+Standard & Base & \textbf{22.0}
& \textbf{32.1}
& 36.9 
& \textbf{32.3} \\ \bottomrule
\end{tabular}%
} 
\vspace{-2mm}
\caption{Open-vocabulary object detection results on LVIS dataset. Without sophisticated training strategy, our detector still outperforms ViLD* on most metrics. Using same training strategy, our open-vocabulary detector beats the fully-supervised Mask RCNN for all metrics.} 
\label{tab:zeroshot-concurrent}
\end{table*}

\smallskip
\noindent \textbf{Evaluation protocol and metrics}. We adopt the standard object detection metrics: mean Average Precision (AP) and AP50 (AP at an intersection over union of 0.5). We evaluate our models on two benchmarks for open-vocabulary object detection, including COCO and LVIS. On COCO, we report AP50 and follow the evaluation settings in \cite{zareian2021open}: (1) only predicting and evaluating novel categories (Novel), (2) only predicting and evaluating base categories (Base), (3) a generalized setting that predicts and evaluates all categories (Generalized). On LVIS, we follow the benchmark of \cite{gu2021zero} where the rare objects are defined as novel categories. We report AP for novel categories (APr), base categories (APc, APf) and all categories (mAP), respectively.

\smallskip
\noindent \textbf{Implementation details}.
\textit{During pretraining}, the default student model and teacher model were both ResNet50~\cite{he2016deep} of pretrained CLIP. The RPN used in pretraining was trained with the base categories of LVIS dataset. Our default model was pretrained on CC3M dataset with the concepts parsed from COCO Cap. SGD was used with the image batch of 96, initial learning rate of 0.002, maximum iteration of 600k, and 100 regions per image. 
\textit{For transfer learning} of object detection, our detectors were developed on Detectron2~\cite{wu2019detectron2} using Faster RCNN~\cite{faster-rcnn} with ResNet50-C4 architecture. The RPN used in transfer learning was trained by the base categories of target dataset (\eg, the transfer learning on COCO used the RPN trained on COCO). SGD was used with image batch of 16, initial learning rate 0.002, and 1x schedule. The weight of background category was set to 0.2/0.8 on COCO/LVIS. Focal scaling was particularly applied to COCO training with $\gamma$ as 0.5. 
\textit{For zero-shot inference} of object detection, RPN was the same as pretraining stage and NMS threshold was set to 0.9. For all experiments, the temperature $\tau$ was 0.01.

\subsection{Transfer Learning for Object Detection}
We present the results of transfer learning for open-vocabulary object detection on COCO and LVIS datasets. Additionally, we report results for fully supervised setting where all categories are used during training.

\begin{table*}[]
\centering
\resizebox{0.81\textwidth}{!}{%
\begin{tabular}{lll|ll|cc|cccc}
\toprule
\multicolumn{3}{c|}{\multirow{2}{*}{\begin{tabular}[c]{@{}c@{}}Visual Encoder Pretraining\end{tabular}}} & \multicolumn{2}{c|}{\multirow{2}{*}{\begin{tabular}[c]{@{}c@{}}Detector Training\end{tabular}}} & \multicolumn{2}{c|}{COCO} & \multicolumn{4}{c}{LVIS} \\
\multicolumn{3}{c|}{} & \multicolumn{2}{c|}{} & \multicolumn{2}{c|}{Train: 80, Test: 80} & \multicolumn{4}{c}{Train: 1203, Test: 1203} \\ 
Method & Dataset & Backbone & Method & Backbone & AP50 & mAP & APr & APc & APf & mAP \\ \midrule
Cls-ResNet~\cite{he2016deep} & ImageNet & RN50 & FR-CNN~\cite{faster-rcnn} & RN50-C4 & 55.9 & 35.7 & 11.9 & 22.0 & 29.7 & 23.3 \\
CLIP~\cite{radford2021learning} & CLIP400M & RN50 & Ours & RN50-C4 & 56.3 & 36.4 & 16.0 & 25.0 & 32.0 & 26.2 \\
Ours & CC3M & RN50 & Ours & RN50-C4 & {59.8} & {38.8} & {18.6} & {27.8} & {34.8} & {29.0} \\ 
Ours & CC3M & RN50x4 & Ours & RN50x4-C4 & \textbf{64.4} & \textbf{42.7} & \textbf{24.5} & \textbf{32.0} & \textbf{36.5} & \textbf{32.5} \\ 
\bottomrule
\end{tabular}%
} 
\vspace{-2mm}
\caption{Fully supervised object detection results on COCO and LVIS datasets. Our detector initialized by our pretrained visual encoder converges faster and significantly outperforms the petrained backbones of ImageNet and CLIP on all metrics at 1x schedule.}
\label{tab:fully-supervised}
\end{table*}

\begin{table*}[]
\centering
\resizebox{0.70\textwidth}{!}{%
\begin{tabular}{lll|c|ccc|cccc}
\toprule
\multicolumn{3}{c|}{Visual Encoder Pretraining} & \multirow{2}{*}{\begin{tabular}[c]{@{}c@{}}Region \\ Proposals\end{tabular}} & \multicolumn{3}{c|}{COCO} & \multicolumn{4}{c}{LVIS} \\ 
Method & Dataset & Backbone &  & Novel & Base & All & APr & APc & APf & mAP \\ \midrule
OVR~\cite{zareian2021open} & COCO Cap & RN50 & GT & 46.7 & \multicolumn{1}{l}{43.7} & \multicolumn{1}{l|}{44.5} & - & - & - & - \\
CLIP~\cite{radford2021learning} & CLIP400M & RN50 & GT & 58.6 & 58.2 & 58.3 & 40.3 & 41.7 & 43.6 & 42.2 \\
Ours & CC3M & RN50 & GT & {60.5} & {61.7} & {61.4} & {40.7} & {43.5} & {47.0} & {44.4} \\ 
Ours & CC3M & RN50x4 & GT & \textbf{65.2} & \textbf{65.6} & \textbf{65.5} & \textbf{50.1} & \textbf{50.1} & \textbf{51.7} & \textbf{50.7} \\ \midrule
OVR~\cite{zareian2021open} & COCO Cap & RN50 & RPN & 24.6 & 17.9 & 19.6 & - & - & - & - \\
CLIP~\cite{radford2021learning} & CLIP400M & RN50 & RPN & 29.7 & 24.0 & 25.5 & {11.6} & 9.6 & 7.6 & 9.2 \\
Ours & CC3M & RN50 & RPN & {31.4} & {25.2} & {26.8} & 10.9 & {10.4} & {8.2} & {9.6} \\ 
Ours & CC3M & RN50x4 & RPN & \textbf{34.6} & \textbf{27.9} & \textbf{29.6} & \textbf{13.8} & \textbf{12.1} & \textbf{9.4} & \textbf{11.3} \\\bottomrule
\end{tabular}%
}  
\vspace{-2mm}
\caption{Zero-shot inference with ground-truth (GT) boxes or RPN boxes on COCO and LVIS dataset. All models use RoIAlign to extract visual representation of proposed image regions. Our pretrained model outperforms baselines by a clear margin across datasets.
} 
\label{tab:zeroshot-inference}
\end{table*}

\subsubsection{Open-Vocabulary Object Detection}

\noindent \textbf{Setup}. The detectors are trained by base categories while evaluated on base and novel categories (\eg, 48/866 base categories and 17/337 novel categories on COCO/LVIS). To compare with ViLD~\cite{gu2021zero}, all experiments on LVIS additionally use mask annotation to train detector.

\smallskip
\noindent \textbf{Baselines}. We consider several baselines as follows:
\squishlist
\item \textbf{Zero-shot object detectors} (SB~\cite{bansal2018zero}, DELO~\cite{Zhu_2020_CVPR}, PL~\cite{Rahman_aaai2020}): Zero-shot object detection is the closest area to open-vocabulary object detection. These detectors usually rely on the pretrained word embeddings of object classes for generalization to novel categories. 
\item \textbf{Open-vocabulary object detectors} (OVR~\cite{zareian2021open}, ViLD~\cite{gu2021zero}): These detectors leverage pretrained vision-language models that have learned a large vocabulary from image-text pairs. OVR is our close competitor in the sense that we both pretrain visual encoders and use them as the detector initialization. ViLD is a recent unpublished work that focuses on detector training by distilling visual features of a pretrained model from CLIP. ViLD specially uses the data augmentation of copy-paste~\cite{ghiasi2021simple} with 16x training schedule. 
\item \textbf{Fully supervised detectors}: On COCO, we include the supervised baseline from OVR which is a Faster RCNN~\cite{faster-rcnn} trained by the base categories with 1x schedule. On LVIS, we include the supervised baseline from ViLD which is a Mask RCNN~\cite{he2017mask} trained by base and novel categories with special data augmentation as ViLD. We additionally report a Mask RCNN trained in standard 1x schedule from Detectron2~\cite{wu2019detectron2}. 
\item \textbf{Our detector variants}: We consider initializing our detector with different pretrained visual encoders, including CLIP and our model pretrained on COCO Cap.
\squishend

\noindent \textbf{Results}. Table~\ref{tab:zeroshot-main} and Table~\ref{tab:zeroshot-concurrent} show the results on COCO and LVIS datasets, respectively.

On COCO dataset, initialized by our pretrained backbone, our detector significantly outperforms previous published SoTA method OVR~\cite{zareian2021open} on all metrics (\eg, 31.4 vs.\ 22.8 on novel categories). Compared with the CLIP backbone from which we start our region-based pretraining, our model brings a remarkable gain across all metrics, particularly +17.2 AP50 on novel categories. When compared with ViLD, an unpublished SoTA method with sophisticated training strategy, our model is still comparable on Base and All, while substantially better on Novel (\eg, 31.4 vs.\ 27.6) which is the main focus in open-vocabulary detection. 
On LVIS dataset, with comparable backbone size (RN50x4-C4 of ours: 83.4M, RN152-FPN of ViLD: 84.1M), our detector outperforms ViLD by a large margin (\eg, +2.2 APr and +3.6 mAP). Note that these superior detection results on COCO and LVIS are achieved by using a single pretrained backbone, with standard data augmentation and 1x training schedule. These results suggest that our region-based vision-language pretraining has learned better alignment between image regions and object concepts, and thus facilitates open-vocabulary object detection.

\subsubsection{Fully Supervised Object Detection}

\noindent \textbf{Setup}. Detection annotation of all object categories are used during training and evaluation. Again, all experiments on LVIS additionally use mask annotation to train detector.

\smallskip
\noindent \textbf{Baselines}. We consider the following baselines: (1) Faster RCNN~\cite{faster-rcnn} intialized by ImageNet pretrained backbone: This is a common object detector in the community. (2) Our detector initialized by pretrained CLIP. This baseline is to validate our proposed pretraining method.

\smallskip
\noindent \textbf{Results}. In Table~\ref{tab:fully-supervised}, the detector initialized by our pretrained visual backbone largely outperforms the baselines that are initialized by ImageNet and CLIP backbones (\eg, +2.4 mAP on COCO and +2.8 mAP on LVIS). These results suggest that our proposed pretraining method helps the fully supervised detector converge faster and achieves better performance at 1x schedule. 
Again, when using RN50x4 as the backbone for both teacher model and student model, the performance is significantly improved (eg, 38.8 vs.\ 42.7 mAP on COCO, 29.0 vs.\ 32.5 on LVIS).

\subsection{Zero-shot Inference for Object Detection}

\noindent \textbf{Setup}. 
Without finetuning on the detection annotation, the pretrained vision-language models are directly used to recognize the proposed regions. We use the same evaluation datasets and metrics as the experiments in transfer learning. We consider two types of region proposals: (1) The ground-truth bounding boxes are used as region proposals. This setting aims at evaluating the recognition performance by eliminating the localization error. (2) The region proposals come from a RPN which is also used in pretraining. The performance in this setting is dependent on both the quality of RPN and recognition ability.

\smallskip
\noindent \textbf{Baselines}. We consider two baselines: (1) OVR~\cite{zareian2021open} pretrains visual backbone on image-text pairs of COCO Cap which has close object concepts as COCO detection dataset. We evaluate the pretrained model provided in their code base.  (2) CLIP~\cite{radford2021learning} is pretrained on 400M image-text pairs. Both OVR and CLIP pretrain model on the image-text pairs while our pretraining leverages the created region-text pairs for learning visual region representation. 

\smallskip
\noindent \textbf{Results}. Table~\ref{tab:zeroshot-inference} summarizes the results. With ideal region proposals, our pretrained model outperforms CLIP baseline by a clear margin across datasets (\eg, 61.4 vs.\ 58.3 All AP50 on COCO, 44.4 vs.\ 42.2 mAP on LVIS). When compared with OVR, our model demonstrates a much larger margin (\eg, 61.4 vs.\ 44.5 All AP50 on COCO), not to mention that OVR is pretrained on the same dataset as evaluation. Even if using RPN proposals, our model still clearly outperforms CLIP and OVR (\eg, 26.8 vs.\ 19.6 \& 25.5 on COCO, 9.6 vs.\ 9.2 on LVIS). These promising results suggest that our pretraining method with region-text alignment improves the visual recognition ability for image regions. 
With RN50x4 architecture as the backbones of teacher and student models, the zero-shot inference performance is further improved across datasets and different types of region proposals (\eg, +6.3 mAP on LVIS with GT boxes, +2.8 All on COCO with RPN boxes).

\begin{figure*}
	\centering
	\includegraphics[width=0.95\linewidth]{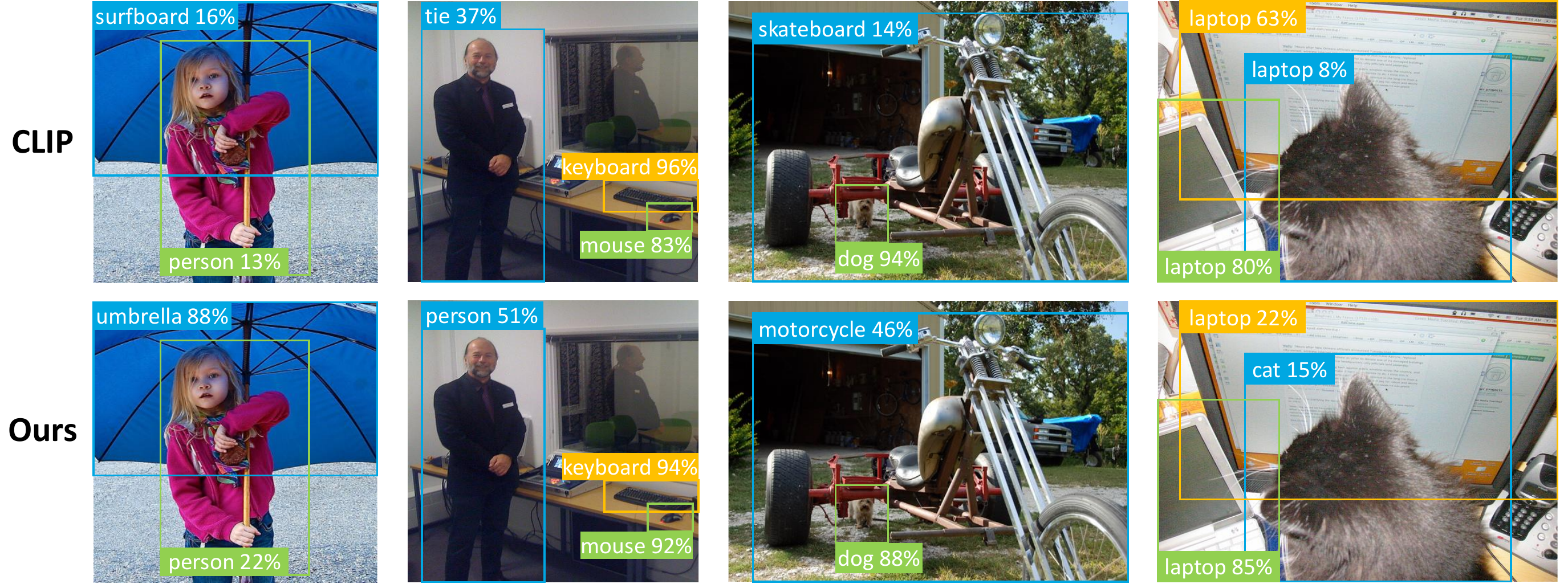} 
	\vspace{-2mm}
    \caption{Visualization of zero-shot inference on COCO dataset with {\it ground-truth boxes}. Without finetuning, the pretrained models (top: CLIP, bottom: Ours) are directly used to recognize image regions into 65 categories in COCO. (Image IDs: 9448, 9483, 7386, 4795)} 
    \vspace{+2mm}
    \label{fig:visualization}
\end{figure*}

\begin{table}[]
\centering
\resizebox{0.47\textwidth}{!}{%
\begin{tabular}{c|c|cc|ccc}
\toprule
\multirow{3}{*}{\begin{tabular}[c]{@{}c@{}}  Region-text \\ Pairs \end{tabular}} & \multirow{3}{*}{\begin{tabular}[c]{@{}c@{}} Image-text \\ Pairs \end{tabular}} & \multicolumn{2}{c|}{\multirow{2}{*}{\begin{tabular}[c]{@{}c@{}}COCO\\ Zero-shot Inference\end{tabular}}} & \multicolumn{3}{c}{COCO} \\
 &  & \multicolumn{2}{c|}{} & \multicolumn{3}{c}{Generalized (17+48)} \\ 
 &  & All (RPN) & All (GT) & Novel & Base & All \\ \midrule
\checkmark &  & 26.7 & 60.4 & 21.4 & 55.5 & 46.6 \\
\checkmark & \checkmark & 28.0 & 62.8 & 26.8 & 54.8 & 47.5 \\ \bottomrule
\end{tabular}%
}  
\vspace{-2mm}
\caption{Ablation study on pretraining supervision. All models are pretrained on COCO Cap.}
\label{tab:image-region}
\end{table}

\begin{table}[]
\centering
\resizebox{0.42\textwidth}{!}{%
\begin{tabular}{cc|cc|ccc}
\toprule
\multicolumn{2}{c|}{\multirow{2}{*}{\begin{tabular}[c]{@{}c@{}}Region Type \end{tabular}}} & \multicolumn{2}{c|}{\multirow{2}{*}{\begin{tabular}[c]{@{}c@{}}COCO\\ Zero-shot Inference\end{tabular}}} & \multicolumn{3}{c}{COCO} \\
\multicolumn{2}{c|}{} & \multicolumn{2}{c|}{} & \multicolumn{3}{c}{Generalized (17+48)} \\ 
Random & RPN & All (RPN) & All (GT) & Novel & Base & All \\ \midrule
\checkmark &  & 27.1 & 60.8 & 25.2 & 54.5 & 46.9 \\
 & \checkmark & 28.0 & 62.8 & 26.8 & 54.8 & 47.5 \\ \bottomrule
\end{tabular}%
}  
\vspace{-2mm}
\caption{Ablation study on the type of regions used during pretraining. All models are pretrained on COCO Cap.} 
\label{tab:ablation-region}
\end{table}

\subsection{Ablation Study}
The evaluation in this section uses COCO dataset and the same metrics as zero-shot inference and transfer learning.

\smallskip
\noindent \textbf{Pretraining supervision}. Table~\ref{tab:image-region} studies the effect of different pretraining supervisions. Accordingly, though using the region-text pairs already attains plausible results, the additional supervision from image-text pairs can further improve the performance (\eg, +2.4 AP50 with GT boxes on zero-shot inference, +5.4 Novel AP50 on transfer learning). We suspect that image-text pairs provide extra contextual information from global image description which compensates our created region descriptions.

\smallskip
\noindent \textbf{Types of image regions}. Table~\ref{tab:ablation-region} studies the effects of region proposal quality during pretraining. We replace the RPN proposals by sampling the same number of image regions with random location and random aspect ratio. Random boxes hurt zero-shot inference (-2.0 AP50 with GT boxes) while reserve comparable performance in transfer learning (46.9 vs.\ 47.5 All AP50). These results indicate that our pretraining is robust to the quality of region proposals. Zero-shot inference benefits from higher quality of proposals but the gap becomes smaller when human supervision is available to finetune the model.

\smallskip
\noindent \textbf{Pretraining dataset and concept pool}. In Table~\ref{tab:ablation-concept}, using COCO Cap dataset or using the COCO concepts achieves better zero-shot inference performance (62.8 vs.\ 61.4 vs.\ 60.8 AP50 with GT boxes). We hypothesize that COCO Cap has a smaller domain gap to COCO detection dataset. However, the model pretrained on CC3M achieves significant boost on transfer learning (47.5 vs.\ 50.4 All AP50). We conjecture that the model learns more generic visual representation from a larger number of images in CC3M.

\smallskip
\noindent \textbf{Pretraining losses}. Table~\ref{tab:ablation-loss} studies the effects of different losses. With both contrastive loss and distillation loss, the model achieves close results as distillation-only model on zero-shot inference (\eg, 62.8 vs.\ 63.1 AP50 with GT boxes) while the best performance on transfer learning (\eg, 26.8 Novel AP50). These results suggest that two losses play different roles. Distillation loss helps to inherit the visual-semantic knowledge from the teacher model, while contrastive loss enforces more discriminative representations for transfer learning.

\begin{table}[]
\centering
\resizebox{0.465\textwidth}{!}{%
\begin{tabular}{c|c|cc|ccc}
\toprule
\multirow{3}{*}{\begin{tabular}[c]{@{}c@{}}Pretraining\\ Dataset\end{tabular}} & \multirow{3}{*}{\begin{tabular}[c]{@{}c@{}}Concept\\ Pool Source\end{tabular}} & \multicolumn{2}{c|}{\multirow{2}{*}{\begin{tabular}[c]{@{}c@{}}COCO\\ Zero-shot Inference\end{tabular}}} & \multicolumn{3}{c}{COCO} \\
 &  & \multicolumn{2}{c|}{} & \multicolumn{3}{c}{Generalized (17+48)} \\ 
 &  & All (RPN) & All (GT) & Novel & Base & All \\ \midrule
COCO Cap & COCO Cap & 28.0 & 62.8 & 26.8 & 54.8 & 47.5 \\
CC3M & COCO Cap & 26.8 & 61.4 & 31.4 & 57.1 & 50.4 \\
CC3M & CC3M & 26.5 & 60.8 & 29.1 & 56.0 & 49.0 \\ \bottomrule
\end{tabular}%
}  
\vspace{-2mm}
\caption{Ablation study on the pretraining datasets and the source of concept pool.}
\label{tab:ablation-concept}
\end{table}

\begin{table}[]
\centering
\resizebox{0.46\textwidth}{!}{%
\begin{tabular}{cc|cc|ccc}
\toprule
\multicolumn{2}{c|}{\multirow{2}{*}{\begin{tabular}[c]{@{}c@{}} Pretraining Loss \end{tabular}}} & \multicolumn{2}{c|}{\multirow{2}{*}{\begin{tabular}[c]{@{}c@{}}COCO\\ Zero-shot Inference\end{tabular}}} & \multicolumn{3}{c}{COCO} \\
\multicolumn{2}{c|}{} & \multicolumn{2}{c|}{} & \multicolumn{3}{c}{Generalized (17+48)} \\ 
Contrastive & Distillation & All (RPN) & All (GT) & Novel & Base & All \\ \midrule
\checkmark &  & 25.2 & 58.2 & 21.8 & 54.2 & 45.8 \\
 & \checkmark & 27.8 & 63.1 & 24.1 & 54.6 & 46.7 \\
\checkmark & \checkmark & 28.0 & 62.8 & 26.8 & 54.8 & 47.5 \\ \bottomrule
\end{tabular}%
}  
\vspace{-2mm}
\caption{Ablation study on losses during pretraining. All models use image-level contrastive loss pretrained on COCO Cap.} 
\label{tab:ablation-loss}
\end{table}

\smallskip
\noindent \textbf{Teacher model and student model}. Table~\ref{tab:teacher} studies the effects of using different teacher and student models. Compared with the default setting at first row, using ResNet50x4 as the teacher model can largely improve the zero-shot inference performance (+4.2 AP50 with GT boxes). However, in the transfer learning setting, the performance using a stronger teacher remains roughly the same (both are 50.4 AP50 for All). When we further replace the student model with ResNet50x4, the transfer learning performance is significantly boosted (+5.3 AP50 for All), but the zero-shot inference performance remains (29.6 vs.\ 29.3 AP50 with RPN boxes). Based on these results, we conjecture that zero-shot inference performance relies on the teacher model that guides the region-text alignment, while transfer learning is more likely constrained by the capacity of student model.

\begin{table*}[!ht]
\begin{minipage}{0.65\linewidth}
\setlength{\tabcolsep}{3.8pt}
\centering
\footnotesize
\begin{tabular}{c|c|cc|ccc}
\toprule
\multirow{3}{*}{\begin{tabular}[c]{@{}c@{}}Teacher\\ Backbone\end{tabular}} & \multirow{3}{*}{\begin{tabular}[c]{@{}c@{}}Student\\ Backbone\end{tabular}} & \multicolumn{2}{c|}{\multirow{2}{*}{\begin{tabular}[c]{@{}c@{}}COCO\\ Zero-shot Inference\end{tabular}}} & \multicolumn{3}{c}{COCO} \\
 &  & \multicolumn{2}{c|}{} & \multicolumn{3}{c}{Generalized (17+48)} \\ 
 &  & All (RPN) & All (GT) & Novel & Base & All \\ \midrule
RN50 & RN50 & 26.8 & 61.4 & 31.4 & 57.1 & 50.4 \\
RN50x4 & RN50 & 29.3 & 65.6 & 30.8 & 57.3 & 50.4 \\ 
RN50x4 & RN50x4 & 29.6 & 65.5 & 39.3 & 61.6 & 55.7 \\ \bottomrule
\end{tabular}%
\vspace{-2mm}
\captionof{table}{Ablation study on COCO with different teacher and student models in pretraining. All models are pretrained on CC3M dataset.}
\label{tab:teacher}
\end{minipage}\hfill
\begin{minipage}{0.3\linewidth}
\setlength{\tabcolsep}{2.8pt}
\centering
\footnotesize
\begin{tabular}{c|ccc}
\toprule
\multirow{3}{*}{\begin{tabular}[c]{@{}c@{}}Focal Scaling\end{tabular}} & \multicolumn{3}{c}{COCO} \\
 & \multicolumn{3}{c}{Generalized (17+48)} \\ 
 & Novel & Base & All \\ \midrule
 & 22.6 & 58.5 & 49.1 \\
\checkmark & 31.4 & 57.1 & 50.4 \\ \bottomrule
\end{tabular}%
\vspace{-2mm}
\captionof{table}{Ablation study on effects of focal scaling during transfer learning for object detection.}
\label{tab:ablation-focal}
\end{minipage}
\end{table*}

\smallskip
\noindent \textbf{Focal scaling}. Table~\ref{tab:ablation-focal} studies the effects of focal scaling during transfer learning. With focal scaling, the finetuned detector achieves a better balance between novel categories and base categories on COCO dataset. We conjecture that the detector overfits to the small set of base categories in COCO (\eg, 48 base categories), which hurts the generalization on novel categories. Focal scaling effectively alleviates the potential overfitting.

\subsection{Discussion}

\noindent \textbf{Visualization}. Fig.~\ref{fig:visualization} visualizes the results of zero-shot inference with ground-truth boxes and 65 categories from COCO dataset. Our model predicts more reasonable categories than CLIP (\eg, the blue regions in 1st and 2nd columns are correctly predicted as ``umbrella'' and ``person'' by our model). These results suggest that our proposed region-based vision-language pretraining can help to recognize image regions precisely.

\begin{figure}
	\centering
	\includegraphics[width=0.80\linewidth]{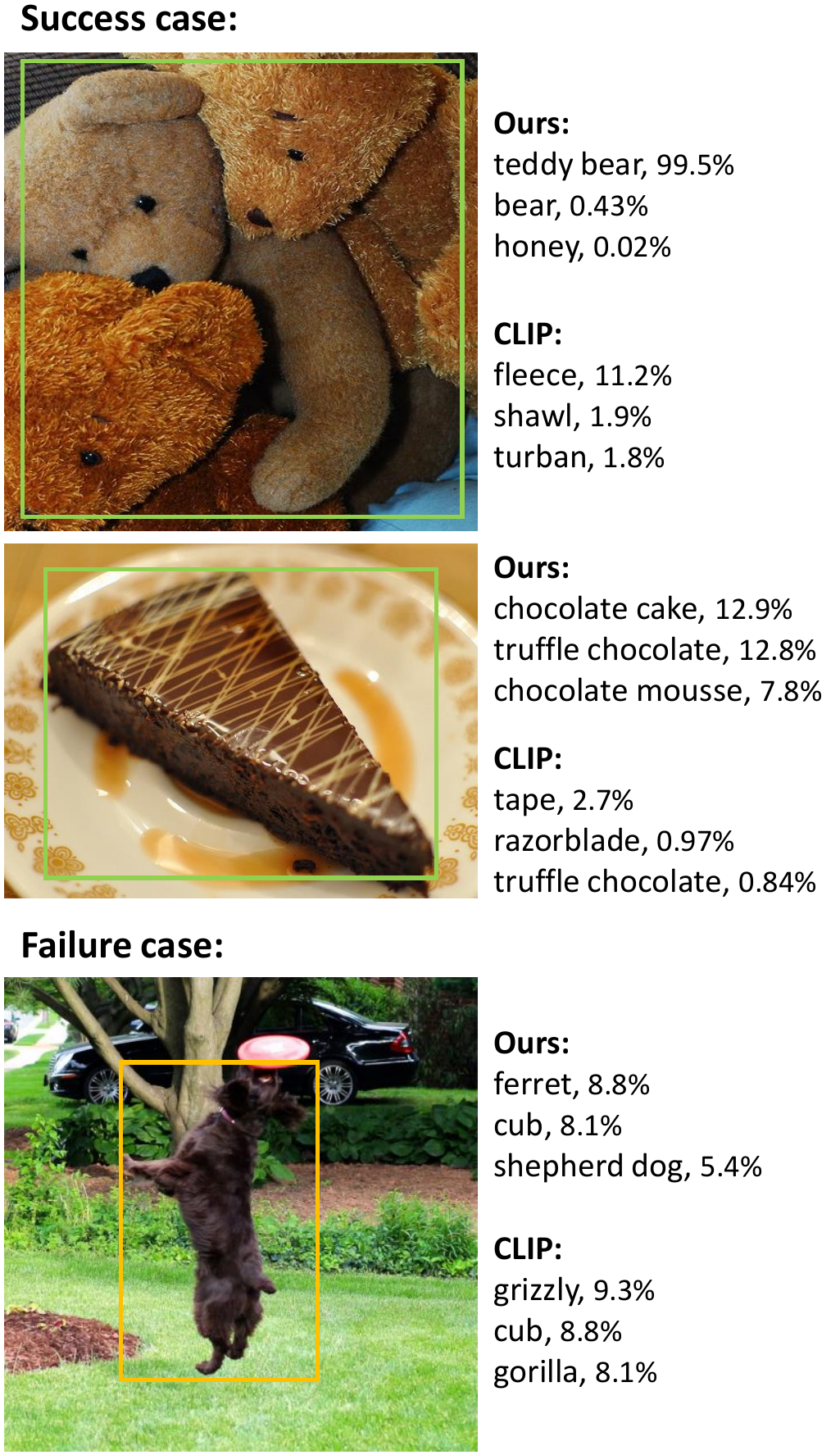} 
    \caption{Visualization of zero-shot inference on COCO dataset with {\it ground-truth boxes}. Without finetuning, the pretrained models are asked to predict 1203 categories from LVIS dataset. We show the top-3 predicted categories from our pretrained model and pretrained CLIP model. (Image IDs: 776, 13597, 17029)} 
    \label{fig:supplement_visualization}
\end{figure}

Further, the pretrained models can predict the customized object concepts by simply replacing the language embeddings of target categories. Fig.~\ref{fig:supplement_visualization} visualizes results of zero-shot inference with ground-truth boxes and 1203 categories from LVIS dataset, instead of the small set of 65 categories from COCO dataset. We show the {\it top-3} predictions for each region with their confidence scores. 

As shown by the successful cases in Fig.~\ref{fig:supplement_visualization}, our pretrained model can correctly recognize the image regions while the CLIP model often fails to predict the correct labels (\eg, ``teddy bear'' is predicted by our model with a high confidence score 99.5\%). Interestingly, other than the most-confident category, our model can also predict reasonable categories with top-3 scores (\eg, ``bear'' in 1st example and ``truffle chocolate'' in 2nd example). Even in the failure case where both CLIP and our model fail to recognize the dog as most-confident category, our model can still recognize the image region as visually similar concepts (\eg, ``ferret'' and ``cub'') or a fine-grained type of dog (\eg, ``shepherd dog''). On the contrary, CLIP predicts less visually similar concepts, such as ``grizzly'' and ``gorilla''.

\smallskip
\noindent \textbf{Limitations}. Our work has several limitations that can be further investigated.
(1) We focus on learning the object concepts without explicitly attending to other information in natural language, such as object attributes and object relationships, which are beneficial to some vision tasks (\eg, visual grounding). Learning comprehensive region representations can be a future work. 
(2) Our method relies on CLIP's visual-semantic space and has not updated the language encoder. When given similar scale of data as CLIP, unfreezing the language encoder may bring more gain in our region-based language-image pretraining.

\section{Conclusion}

In this paper, we proposed a novel region-based vision-language pretraining method that learned to match image regions and their descriptions. Our key innovation is a scalable approach to associate region-text pairs beyond the tokens presented in the paired text data without using human annotation. Learning from such region-level alignment, our pretrained model established new state of the art when transferred to open-vocabulary object detection on COCO and LVIS datasets. Moreover, our pretrained model demonstrated promising results on zero-shot inference for object detection. We hope that our work can shed light on vision-language pretraining for visual region understanding.

{\small
\bibliographystyle{ieee_fullname}
\bibliography{egbib}
}

\end{document}